\documentclass[conference]{IEEEtran}
\IEEEoverridecommandlockouts
\usepackage{cite}
\usepackage{amsmath,amssymb,amsfonts}
\usepackage{algorithmic}
\usepackage{graphicx}
\usepackage{textcomp}
\usepackage{xcolor}

\usepackage{hyperref}
\usepackage{booktabs}  
\usepackage{multirow}  
\usepackage{xcolor}    
\usepackage{colortbl}  
\usepackage{amsmath}   
\usepackage{multicol}

\usepackage{subcaption}
\usepackage{pifont}
\usepackage{tcolorbox}

\usepackage[ruled,vlined]{algorithm2e}
\usepackage{amsmath}
\usepackage{amssymb}

\usepackage{listings}
\usepackage{xcolor}
\definecolor{mydarkgreen}{rgb}{0.0, 0.7, 0.0} 
\definecolor{myred}{rgb}{0.7, 0.0, 0.0}
\def\BibTeX{{\rm B\kern-.05em{\sc i\kern-.025em b}\kern-.08em
    T\kern-.1667em\lower.7ex\hbox{E}\kern-.125emX}}
\begin{document}

\title{RSL-SQL: Robust Schema Linking in Text-to-SQL Generation}

\author{
  Zhenbiao Cao\textsuperscript{1},
  Yuanlei Zheng\textsuperscript{1},
  Zhihao Fan\textsuperscript{2}, \\
  Xiaojin Zhang\textsuperscript{1}, 
  Wei Chen\textsuperscript{1\dag} and 
  Xiang Bai\textsuperscript{1} 
  \vspace{0.5em}  \\   
  \textsuperscript{1}Huazhong University of Science and Technology, \textsuperscript{2}Alibaba Inc. \\
  \textsuperscript{\dag}\texttt{lemuria\_chen@hust.edu.cn} \\
}


\maketitle

\begin{abstract}
Text-to-SQL generation aims to translate natural language questions into SQL statements. In Text-to-SQL based on large language models, schema linking is a widely adopted strategy to streamline the input for LLMs by selecting only relevant schema elements, therefore reducing noise and computational overhead. However, schema linking faces risks that require caution, including the potential omission of necessary elements and disruption of database structural integrity. To address these challenges, we propose a novel framework called RSL-SQL that combines bidirectional schema linking, contextual information augmentation, binary selection strategy, and multi-turn self-correction. We improve the recall of pattern linking using forward and backward pruning methods, achieving a strict recall of 94\% while reducing the number of input columns by 83\%. Furthermore, it hedges the risk by voting between a full mode and a simplified mode enhanced with contextual information. Experiments on the BIRD and Spider benchmarks demonstrate that our approach achieves SOTA execution accuracy among open-source solutions, with \textbf{67.2\%} on BIRD and \textbf{87.9\%} on Spider using GPT-4o. Furthermore, our approach outperforms a series of GPT-4 based Text-to-SQL systems when adopting DeepSeek (much cheaper) with same intact prompts. Extensive analysis and ablation studies confirm the effectiveness of each component in our framework. The codes are available at \href{https://github.com/Laqcce-cao/RSL-SQL}{https://github.com/Laqcce-cao/RSL-SQL.}
\end{abstract}


\section{Introduction}

The task of translating natural language questions into structured query language (SQL), known as \emph{Text-to-SQL} or \emph{Text2SQL} generation, is crucial for enabling non-expert users to interact with relational databases~\cite{wang2019rat,qin2022survey}. By bridging the gap between natural language and structured query languages, Text-to-SQL systems democratize data access and empower users to extract valuable insights without deep technical knowledge~\cite{katsogiannis2023survey}. 


In very recent years, leveraging the powerful comprehension and generation capabilities of Large Language Models (LLMs)~\cite{achiam2023gpt,openai2024gpt4oannouncement,anthropic2024claude} for Text-to-SQL tasks has become a primary approach for boosting performance, and prompt engineering has emerged as the mainstream technical strategy. A typical prompt provided to the LLM for Text-to-SQL usually includes a description of the database, user queries, and few-shot demonstrations~\cite{gao2023text,pourreza2024din}, which allows the system to be applicable to various databases. Intuitively, assuming the LLM possesses sufficiently strong capabilities, the more precise and detailed the database description, the better the quality of the generated SQL queries. Features such as the database's structure, annotations, sample data, and relational mappings have been shown to improve Text-to-SQL performance in specific scenarios~\cite{li2024pet,lee2024mcs}. 


Fine-grained descriptions of databases present challenges. It is common for databases, especially large-scale industrial databases, to have hundreds or thousands of fields. Incorporating complete database features in the prompt leads to excessive input tokens, increased computational costs, and, critically, the introduction of substantial noise~\cite{talaei2024chess}. Since user queries typically refer to a small proportion of database schema elements, a large amount of irrelevant schema information can confuse LLM and degrade performance. To mitigate this, \emph{schema linking}~\cite{lei2020re,pourreza2024din} techniques have been widely adopted to identify and include only relevant schema elements in the prompts. 



However, schema linking introduces both opportunities and risks. On the one hand, pruning irrelevant schema elements can reduce input complexity and focus the LLM's attention on pertinent information, which can potentially improve SQL generation accuracy. On the other hand, during the simplification process, certain tables or columns that are important but not immediately obvious in the original problem may be overlooked, leading to missing necessary information in the generated SQL. This results in \textbf{\emph{Risk 1) If schema linking fails to identify all necessary tables and columns, the generated SQL will inevitably be erroneous}} (assuming that the LLM does not generate hallucinations, meaning the schema elements in the SQL are entirely derived from the input database schema). Additionally, schema linking may overlook relationships between tables (such as foreign key constraints), causing issues in correctly establishing joins in multi-table queries. Moreover, the simplified schema may result in more generic column or table names, amplifying semantic ambiguities. This leads to \textbf{\emph{Risk 2) Recent studies suggest that even when all required elements are identified, schema linking can still cause performance degradation.}} These risks point to a delicate balance: while schema linking can reduce noise and computational cost, it may also remove critical information or context necessary for accurate SQL generation.


Our analysis further emphasizes this trade-off. Simplifying the complete schema through schema linking has dual effects: it can correct queries that were originally incorrect in the complete schema (positive gain, denoted as $y$) but may also cause initially correct queries to become incorrect (negative impact, denoted as $x$). Thus, the net benefit of schema linking depends on the balance between these effects. Schema linking yields a net positive effect only when $y - x > 0$; otherwise, it may introduce more harm than benefit.

To enhance the benefits of schema linking while mitigating its risks, we propose \textbf{RSL-SQL}, a \textbf{R}obust \textbf{S}chema \textbf{L}inking based Text-to-\textbf{SQL} generation framework. Our approach is designed to maximize the positive gain ($y$) and minimize the negative impact ($x$) of schema linking.

In our framework, we first generate a preliminary SQL using complete database schema and achieve high recall rate through \textbf{bidirectional schema linking} to reduce the risk of incomplete recall of schema elements. Next, we simplify the database schema and enhance it with \textbf{rich contextual information} to generate another SQL, which increases the number of corrected queries but also, to a smaller extent, increases the number of newly introduced errors. Subsequently, a \textbf{binary selection strategy} (selecting the better SQL generated from the complete or simplified database schema) is employed to further boost the number of corrected queries while reducing the newly introduced errors. Finally, we employ a \textbf{multi-turn self-correction} approach, integrating feedback from SQL execution results to iteratively refine and optimize bad SQL statements. Through these strategies, RSL-SQL enhances the positive effects of schema linking while mitigating its negative impacts, leading to significant overall performance improvements.

In our experiments, we evaluate the proposed method, RSL-SQL, on the BIRD and Spider datasets, comparing its performance against a range of existing Text-to-SQL methods. The experimental results show that when using GPT-4o as the backbone LLM, RSL-SQL achieves \textbf{67.21\%} execution accuracy and \textbf{70.32\%} valid efficiency score on the BIRD dev dataset (setting a new state-of-the-art for open-source methods), and \textbf{87.9\%} execution accuracy on the Spider test dataset (comparable to state-of-the-art). Moreover, we demonstrate that when using the significantly more cost-effective DeepSeek, RSL-SQL outperforms many GPT-4-based methods on both the BIRD and Spider datasets, despite the per-token cost of GPT-4 being \textbf{215} times higher than DeepSeek. The ablation study reveals that each component of our method contributes to the overall performance gains. Notably, our bidirectional schema linking technique achieves a high strict recall rate of \textbf{94\%} (a new state-of-the-art) on the BIRD dataset, while significantly reducing the average number of columns per query. Both contextual information augmentation and the binary selection strategy are shown to steadily improve accuracy, thereby mitigating the potential risks associated with schema linking.

The \textbf{main contributions} of this paper can be summarized as follows:

\textbf{(1)} We investigate the potential risks associated with schema linking and propose RSL-SQL, a novel Text-to-SQL generation framework with robust schema linking that maximize the benefits while mitigating the risks associated with schema linking, achieving state-of-the-art performance on the BIRD dataset.

\textbf{(2)} Extensive experimental results demonstrate the effectiveness and robustness of the proposed method. Our framework also exhibits good transferability, with its performance surpassing many GPT-4-based methods when using much cheaper DeepSeek, demonstrating excellent cost-effectiveness.

\begin{figure*}[htbp]
    \centering
    \includegraphics[width=1\linewidth]{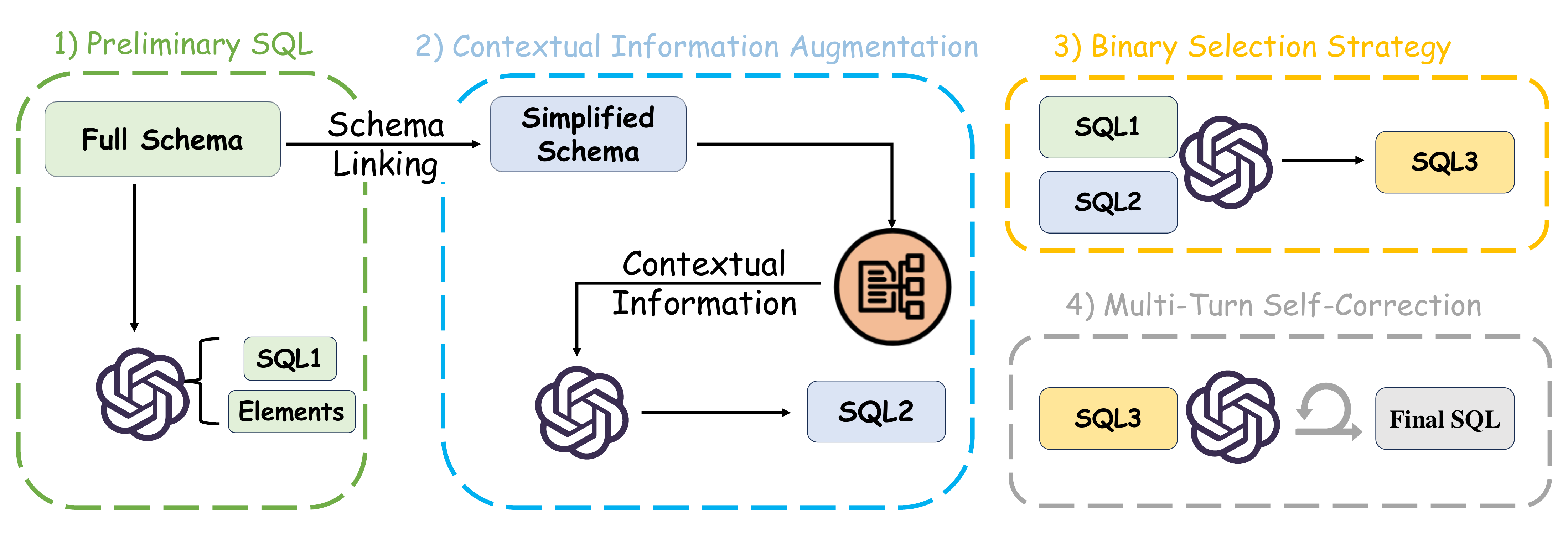}
    \caption{The framework consists of four main components: (1) Preliminary SQL Generation, (2)  SQL Generation with Contextual Information Augmentation, (3)  Binary Selection Strategy, and (4)  Multi-Turn Self-Correction. }
    \label{framework}
\end{figure*}

\section{Related Work}

Early research on Text-to-SQL tasks relies primarily on hand-designed templates \cite{zelle1996learning}. Although these methods perform well in simple scenarios, they are highly dependent on manual rules and are difficult to extend to more complex and diverse application scenarios. With the widespread application of Transformer models, especially models based on sequence-to-sequence architecture \cite{vaswani2017attention,sutskever2014sequence}, text-to-SQL research makes significant progress. For example, IRNet \cite{guo2019towards} and RAT-SQL \cite{wang2019rat} adopt a relationship-aware attention mechanism to tightly integrate database schema information with the SQL generation process. After LLM is widely proven to be powerful in NLP tasks, more and more research explores its potential in Text-to-SQL tasks. Methods such as QDecomp \cite{tai2023exploring}, C3 \cite{dong2023c3}, QDMR \cite{wolfson2021weakly}, and DIN-SQL \cite{pourreza2024din} introduce task decomposition and reasoning strategies, such as Chain of Thought (CoT) \cite{wei2022chain}, to gradually improve SQL generation performance. In the SQL generation process, the method of using LLM to generate multiple candidate SQL statements and select the best candidate has been proven to be effective~\cite{chen2021evaluating,li2022competition,ni2023lever}. For example, multiple candidate SQLs can be generated through different prompts, and then the optimal solution can be selected~\cite{pourreza2024chase}. This strategy can undoubtedly improve the accuracy of the generation. In contrast, our method only generates two candidate SQL statements based on the complete schema and simplified schema respectively, thus significantly reducing the use of tokens and significantly improving efficiency.

Schema linking is a critical step in text-to-SQL tasks that aims to identify relevant database tables and columns associated with natural language queries. To enhance the integration of schema information and capture its relationship with the problem, models such as RAT-SQL \cite{guo2019towards}, SchemaGNN \cite{bogin2019representing}, and ShadowGNN \cite{chen2021shadowgnn} adopt a relationship-aware self-attention mechanism. Furthermore, SADGA \cite{cai2021sadga} introduces a novel dual-graph framework designed to interactively encode and aggregate structural information from natural language questions and database schemas. With LLM's outstanding performance in NLP tasks, many studies attempt to apply it to schema linking tasks. For example, CHESS \cite{talaei2024chess} enables more accurate schema linking by retrieving relevant information from database catalogs and database values. Another approach, MCS-SQL\cite{lee2024mcs}, relies on using multiple prompts and sampling multiple responses from LLM, leveraging LLM's large sample library to filter out irrelevant tables and columns to reduce the likelihood of incorrect matches. We use a bidirectional schema linking approach to dig deep into the external knowledge provided by the database and analyze the relationship between the initial SQL statement generated and the elements required to answer the user's question.

\section{Method}

In this section, we introduce RSL-SQL, a proposed framework for generating Text-to-SQL using LLMs. As shown in Fig.~\ref{framework}, RSL-SQL consists of four key components: 1) bidirectional schema linking, 2) contextual information augmentation, 3) binary selection strategy, and 4) multi-turn self-correction.

The framework begins with bidirectional schema linking to recall all necessary database elements as comprehensively as possible, achieving a high recall rate. Next, contextual information augmentation further enhances the positive gain of schema linking. Subsequently, the binary selection strategy is applied to maximize the positive gain of schema linking while reducing any negative effects. Finally, multi-turn self-correction iteratively refines any erroneous SQL generated. Each of these components will be discussed in detail in the following sections. Alg.~\ref{alg:framework} outlines the step-by-step process of this strategy to generate the final SQL output.

\subsection{Prompt Engineering}

Leveraging LLMs for Text-to-SQL task requires careful consideration of prompt engineering that influence the portability to diverse databases and accuracy of generated SQL queries. In this paper, we considers following key input elements: 

\noindent\textbf{Database Schema} ($\mathcal{S}$) \quad Detailed information about the database structure, such as table names, column names, and foreign key relationships. Providing this schema is crucial for helping LLMs understand real-world diverse database architecture and ensuring that the generated queries reference the appropriate tables and columns.

\begin{figure*}[htbp]
    \centering
    \includegraphics[width=0.95\linewidth]{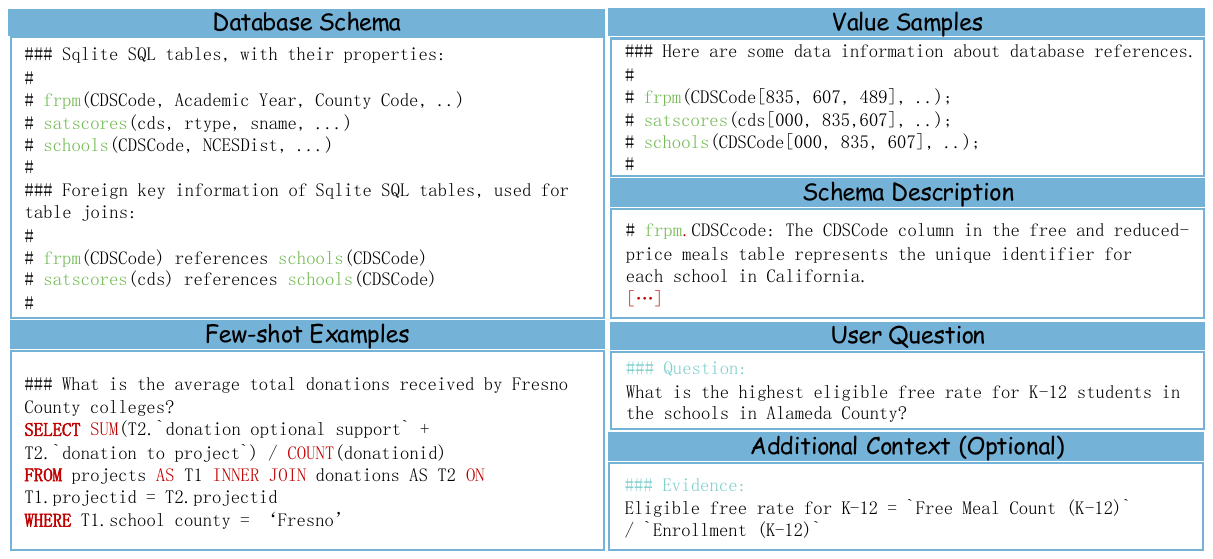}
    \caption{Organization of Elements in the Prompt.}
    \label{prompt_engineer}
\end{figure*}

\noindent\textbf{Value Samples} ($\mathcal{V}$) \quad Randomly selected a few rows from each table, allowing LLMs to gain a better understanding of the potential values of each column and semantics of the tables and columns. For overly long values, we truncate them to maintain a manageable input size.

\noindent\textbf{Schema Descriptions} ($\mathcal{D}$) \quad Textual descriptions that offer additional semantic context about all tables and columns given any database with definite schema. We generate schema descriptions $\mathcal{D}$ utilizing LLMs, following the approach in~\cite{qu2024before}, to enrich semantic context for all schema elements of each database.

\noindent\textbf{Few-Shot Examples} ($\mathcal{E}$) \quad For each question, we retrieve a few pairs of natural language questions and their corresponding SQL queries from the training set. To identify most similar questions to the target question, we employ a Euclidean distance based question selector to select the top-k pairs, exactly following \cite{li2024pet}.

\noindent\textbf{User Question} ($\mathcal{Q}$) \quad Natural language question from the user, serving as the primary input for SQL translation.

\noindent\textbf{Additional Context (Optional)} ($\mathcal{C}$) \quad Any supplementary information that can clarify the user's question, such as definitions, constraints, or domain-specific knowledge. 

We show how each part of above elements is organized in the prompt of LLMs in Fig.~\ref{prompt_engineer}.

\subsection{\textbf{Step 1: Bidirectional Schema Linking (BSL)}}

\begin{figure}[htbp]
    \centering
    \small
    \includegraphics[width=0.95\linewidth]{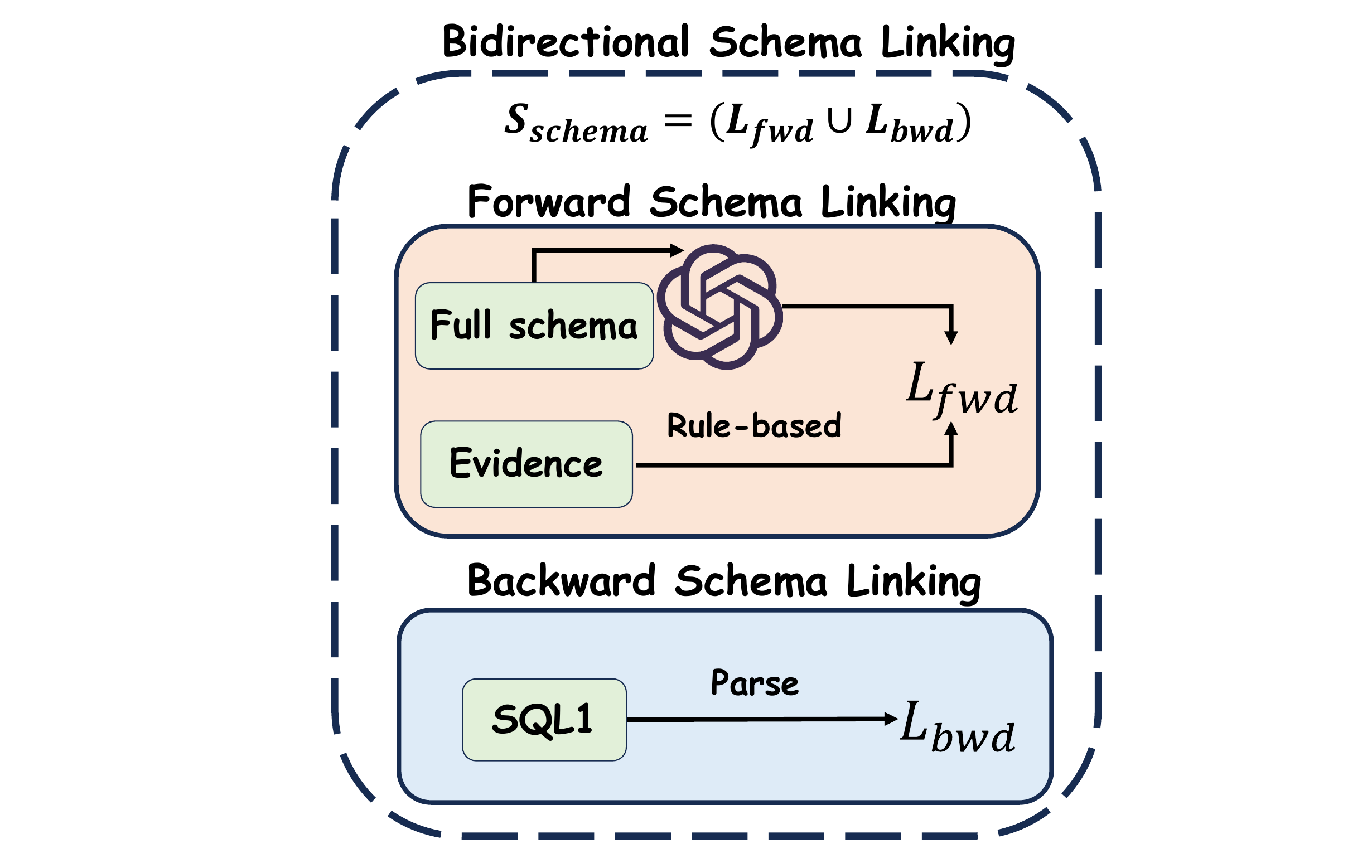}
    \caption{Framework of Bidirectional Schema Linking. \textbf{$L_\text{fwd}$} and \textbf{$L_\text{bwd}$} refer to the results of forward schema linking and backward schema linking respectively.}
    \label{bid}
\end{figure}

We first introduce our bidirectional schema linking approach. The flowchart of bidirectional schema linking is shown in Fig.~\ref{bid}. In this process, a preliminary SQL query is generated based on complete database schema.

\subsubsection{\textbf{Forward Schema Linking (FSL)}} ~ {\emph{Forward Schema Linking} aims to directly identify potentially relevant schema elements  to $\mathcal{Q}$ from the full database schema. Inspired by \cite{lee2024mcs}, we input full schema $\mathcal{S}$, value samples $\mathcal{V}$, and user's question $\mathcal{Q}$. We then prompt LLMs to identify and format relevant tables as \texttt{table\_name} and columns as \texttt{table\_name.column\_name}. If additional context $\mathcal{C}$ is available, we augment this list by traversing all column names and including those that appear in $\mathcal{C}$. This combined list results in a set of tables and columns denoted as $L_{\texttt{fwd}}$.}

\subsubsection{\textbf{Preliminary SQL Generation}} ~ {We aim to generate a preliminary SQL query $\texttt{SQL}_{1}$ using the full schema $\mathcal{S}$. The input components for LLMs include the full schema $\mathcal{S} $, value samples $\mathcal{V}$, few-shot examples $\mathcal{E}$, user's question $\mathcal{Q}$ and additional context $\mathcal{C}$. We also incorporate the results from the forward schema linking, $L_{\texttt{fwd}}$, as supplementary context, which we find can slightly enhance the accuracy of the preliminary SQL query. This can be formalized as: 
\begin{equation}
    \texttt{SQL}_{1} = f_{\text{LLM}}(\mathcal{S}, \mathcal{V}, \mathcal{E}, \mathcal{Q}, \mathcal{C}, L_{\texttt{fwd}})
\end{equation}

We \textbf{do not include} schema descriptions $\mathcal{D}$ since describing each table and column with a paragraph within complete schema would result in overly lengthy input that may exceed the length limit of some LLMs.
}

\subsubsection{\textbf{Backward Schema Linking (BSL)}} ~ {\emph{Backward Schema Linking} is complementary to \emph{Forward Schema Linking}, where we parse the preliminary SQL query $\texttt{SQL}_{1}$ to extract an additional set of referenced tables and columns $L_\texttt{bwd}$. For each element \texttt{table\_name.column\_name} in the database, if \texttt{column\_name} is used in the generated preliminary SQL, it is recalled accordingly. The column name exact match method may cause some redundant columns to be recalled. Of course, you can also use the sqlglot~\cite{sqlglot} tool to parse SQL to accurately recall the \texttt{table\_name.column\_name} used by SQL. 

Both methods have their own advantages and disadvantages: using full column name matching can improve the recall rate, but it may recall more irrelevant columns; while using the SQLGlot tool has the advantage of less noise, but the recall rate will be reduced. In addition, if there is an error in the preliminary SQL itself, if you use the sqlglot tool recall method, problems may still occur even after simplified schema processing. In contrast, although the exact matching method introduces redundant columns, the error probability of SQL generated based on the simplified schema will be reduced. Therefore, we ultimately chose to implement backward schema linking using an exact match of column names.

This step is crucial, as $L_\texttt{bwd}$ should encompass all necessary elements if $\texttt{SQL}_{1}$ is already correct.}

\subsubsection{\textbf{Schema Simplification}} ~ {We merge the results of forward and backward schema linking to obtain a more complete subset of tables and columns, i.e., $L_\texttt{fwd} \cup L_\texttt{bwd}$, and streamline the database schema, value samples, and schema descriptions, resulting in their condensed counterparts $\mathcal{S'}$, $\mathcal{V'}$, and $\mathcal{D'}$, which we called \textbf{Bidirectional Schema Linking}. It can significantly reduce redundant database information irrelevant to the user question, curtail the required input length and reduce the decision space for subsequent SQL generation, while maintaining a high recall for necessary schema elements.}


\subsection{\textbf{Step 2: Contextual Information Augmentation (CIA)}}

\begin{figure}[htbp]
    \centering
    \includegraphics[width=0.95\linewidth]{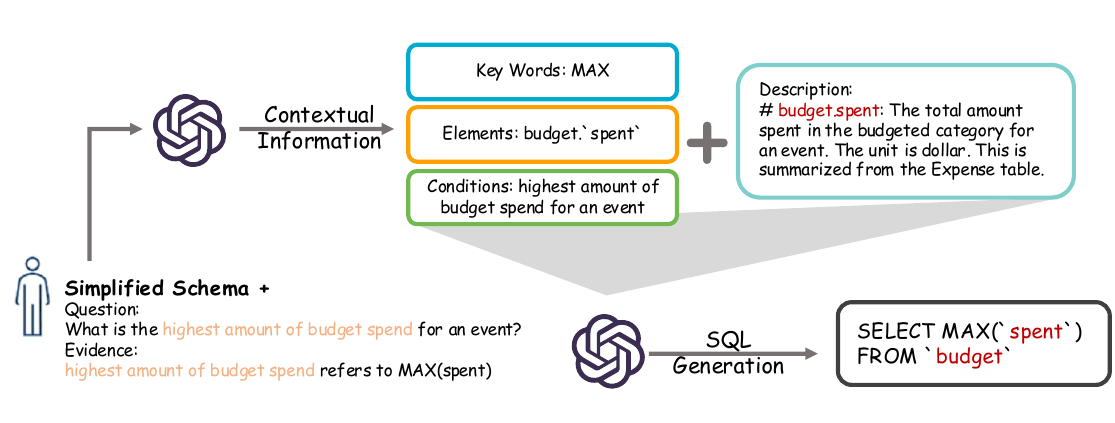}
    \caption{An example of contextual information augmentation.}
    \label{example_cia}
\end{figure}

Schema linking simplifies database structure and focus the LLM’s attention
on pertinent information. To enhance this benefit, we utilize LLM to independently generate key components of an SQL statement, including schema elements, conditions, and keywords contained within the SQL. We then input these components as additional information, along with detailed descriptions of each column in the simplified schema. Finally, we generate the complete SQL statement based on these collective inputs. An example can be seen in Fig.~\ref{example_cia}.

\subsubsection{\textbf{SQL Components Generation}} ~ {We input the user query $\mathcal{Q}$, simplified schema $\mathcal{S'}$, value samples $\mathcal{V'}$, description of columns $\mathcal{D'}$, and optional context $\mathcal{C}$, \emph{SQL Components Generation} aims to to pre-generate each component necessary for SQL generation. This includes:

\begin{itemize}
    \item \textbf{Elements (\( H_\text{E} \))}: A list of tables and columns likely needed in the SQL query, which is exactly the same as forward schema linking. 
    \item \textbf{Conditions (\( H_\text{C} \))}: Possible conditions and constraints for the WHERE clause, through decomposition and analysis of the question. 
    \item \textbf{Keywords (\( H_\text{K} \))}: SQL keywords (e.g., DISTINCT, GRO-UP BY) that may be relevant, by locating key indicator words in the question.
\end{itemize}

We define the simplified schema descriptions $\mathcal{D'}$ along with the LLM-generated \( H_\text{E} \), \( H_\text{C} \), and \( H_\text{K} \) as contextual augmented information, collectively denoted as $H_\text{Aug} = \{\mathcal{D'}, H_\text{E}, H_\text{C}, H_\text{K}\}$, to help LLM to better understand the simplified database schema and target SQL statement.}

\subsubsection{\textbf{SQL Generation with Simplified Schema}} ~ {We generate an another SQL query based on the simplified schema. The prompt for the LLM includes the simplified schema $\mathcal{S'}$, value samples $\mathcal{V'}$, contextual augmented information $H_\text{Aug}$, few-shot examples $\mathcal{E}$, the user question $\mathcal{Q}$ and optional question context $\mathcal{C}$. This process can be formalized as: \begin{equation}
    \texttt{SQL}_{2} = f_{\text{LLM}}(\mathcal{S'}, \mathcal{V'}, H_{\text{Aug}}, \mathcal{E}, \mathcal{Q}, \mathcal{C})
\end{equation}}

\subsection{\textbf{Step 3: Binary Selection Strategy (BSS)}}

\begin{figure}[htbp]
    \centering
    \includegraphics[width=0.95\linewidth]{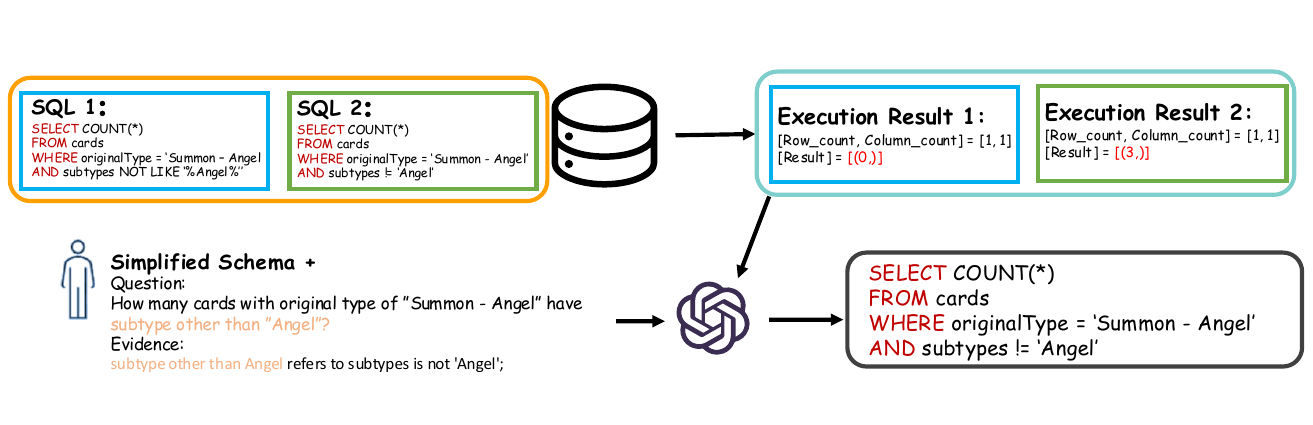}
    \caption{An example of binary selection strategy.}
    \label{example_bss}
\end{figure}

This binary selection strategy is an important step to mitigate the risk of schema linking. In fact, the complete schema and the simplified schema each have their own advantages. The former retains the complete database structure but has redundant information, while the latter removes a lot of redundant information but has the risk of information loss. Using LLM to compare the two independently generated SQLs and select the best one can be more vividly understood as a risk hedging. Fig.~\ref{example_bss} shows a specific example.

We generate two SQL queries: $\texttt{SQL}_{1}$ from the complete schema $\mathcal{S}$ for structural completeness and $\texttt{SQL}_{2}$ from a simplified schema $\mathcal{S'}$ to reduce noise. To select the optimal query $\texttt{SQL}_{3}$, we use a large language model (LLM).

Both $\texttt{SQL}_{1}$ and $\texttt{SQL}_{2}$ are executed, producing results $\mathcal{R}_{1}$ and $\mathcal{R}_{2}$ for correctness evaluation. To assess these queries, we utilize the user's question \( \mathcal{Q} \), additional context \( \mathcal{C} \), simplified schema $\mathcal{S'}$, value samples $\mathcal{V'}$, and few-shot examples $\mathcal{E}$. The LLM then analyzes $\mathcal{R}_{1}$ and $\mathcal{R}_{2}$ to select the optimal query as $\texttt{SQL}_{3}$, ensuring semantic alignment with \( \mathcal{Q} \). This process can be formalized as: \begin{multline}
\texttt{SQL}_{3} = f_{\text{LLM}}(\mathcal{S'}, \mathcal{V'}, \mathcal{D'}, \mathcal{E}, \mathcal{Q}, 
\mathcal{C}, \\ \texttt{SQL}_{1}, \texttt{SQL}_{2}, \mathcal{R}_{1}, \mathcal{R}_{2})
\end{multline}

\subsection{\textbf{Step 4: Multi-Turn Self-Correction (MTSC)}}

Queries that cannot be executed are deemed incorrect. Similarly, empty execution results may indicate faults. We apply rules to identify high-error-probability queries for iterative refinement.


\subsubsection{\textbf{Iterative Refinement Process}} ~ {$\texttt{SQL}_{3}$ may still contain syntax errors or yield empty results. If execution produces a syntax error or an empty result set, we capture and log this outcome as $ E^{(0)}$.

\begin{figure}[htbp]
    \centering
    \includegraphics[width=0.95\linewidth]{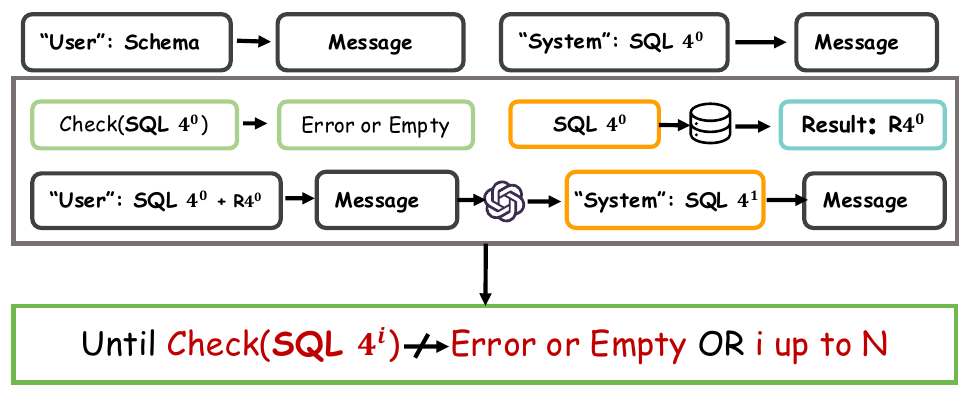}
    \caption{Flowchart of Multi-Turn Self-Correction.}
    \label{example_mtsc}
\end{figure}

For SQL queries with a high likelihood of errors, we employ an multi-turn conservation-based correction process. Initially, as illustrated in Fig.~\ref{example_mtsc}, we provide the user's question \( \mathcal{Q} \), context \( \mathcal{C} \), simplified schema $\mathcal{S'}$, value samples $\mathcal{V'}$, examples $\mathcal{E}$, the query $\texttt{SQL}_{\texttt{4}}^{(0)}$ (initialized as $\texttt{SQL}_{\texttt{3}}$), and the error information $E^{(0)}$ to the LLM. The LLM then generates a new SQL statement, which is executed. If errors or empty results persist, we continue the dialogue by informing the LLM of the SQL error, iterating until reaching the maximum dialogue rounds. This process can be formalized as:\begin{multline}
\texttt{SQL}_{4}^{(i+1)} = f_{\text{LLM}}(\mathcal{S'}, \mathcal{V'}, \mathcal{D'}, \mathcal{E},  \mathcal{Q}, \mathcal{C}, \texttt{SQL}_{4}^{(\leq{i})},E^{(\leq{i})})
\end{multline} 

\subsubsection{\textbf{Termination Condition}} ~ {The refinement process concludes when $\texttt{SQL}_{\texttt{4}}^{i}$ executes successfully without syntax errors, returns a non-empty result, or reaches the maximum number of iterations \( N \).}

\begin{algorithm}[htbp]
\caption{RSL-SQL Framework}
\label{alg:framework}
\KwIn{Full Schema $\mathcal{S}$, Value Samples $\mathcal{V}$, Schema Description $\mathcal{D}$, Few-Shot Examples $\mathcal{E}$, User Question $\mathcal{Q}$, Optional Context $\mathcal{C}$}
\KwOut{Final SQL $\texttt{SQL}_4$}

\SetKwProg{Fn}{Function}{:}{end}

    \Fn{\texttt{$Step_1: BSL$}($\mathcal{E}$, $\mathcal{S}, \mathcal{V}, \mathcal{Q}, \mathcal{C}$)}{
    \tcp{\textbf{Forward Schema Linking:} }
    
    $L_\text{fwd} \gets f_\text{LLM}(\mathcal{S}, \mathcal{V}, \mathcal{Q}, \mathcal{C})$\;
    
    \tcp{\textbf{Preliminary SQL Generation:} }
    
    $\texttt{SQL}_1 \gets f_\text{LLM}(\mathcal{E}, \mathcal{S}, \mathcal{V}, L_\text{fwd}, \mathcal{Q}, \mathcal{C})$\;
    
    \tcp{\textbf{Backward Schema Linking:} }
    
    $L_\text{bwd} \gets f_\text{Rules}(\texttt{SQL}_1, \mathcal{S})$\;
    
    \tcp{\textbf{Schema Simplification:} }
    
    $\mathcal{S'}, \mathcal{V'} \gets \texttt{SimplifySchema}(\mathcal{S}, L_\text{fwd} \cup L_\text{bwd})$\;
    \Return{$\mathcal{S'}, \mathcal{V'}, \texttt{SQL}_1$}
}

\BlankLine
\Fn{\texttt{$Step_2: CIA$}($\mathcal{E}, \mathcal{D'}, \mathcal{S'}, \mathcal{V'}, \mathcal{Q}, \mathcal{C}$)}{
    \tcp{\textbf{Generate SQL Components:} }
    $H_\text{E}, H_\text{C}, H_\text{K} \gets f_\text{LLM}(\mathcal{S'}, \mathcal{V'}, \mathcal{D'}, \mathcal{Q}, \mathcal{C})$\;
    
    \tcp{\textbf{Augment Information:}} 
    $H_\text{Aug} \gets \{\mathcal{D'}, H_\text{E}, H_\text{C}, H_\text{K}\}$\;
    
    \tcp{\textbf{Generate SQL:}} 
    
    $\texttt{SQL}_2 \gets f_\text{LLM}(\mathcal{E}, \mathcal{S'}, \mathcal{V'}, H_\text{Aug}, \mathcal{Q}, \mathcal{C})$\;
    \Return{$\texttt{SQL}_2$}
}
\BlankLine
\Fn{\texttt{$Step_3: BSS$}($\mathcal{E}, \mathcal{S'}, \mathcal{V'}, \mathcal{Q}, \mathcal{C}, \texttt{SQL}_1, \texttt{SQL}_2$)}{

    \tcp{\textbf{Execute $\texttt{SQL}_1$ and $\texttt{SQL}_2:$}}
    Execute $\texttt{SQL}_1$ and $\texttt{SQL}_2$, producing $\mathcal{R}_1$ and $\mathcal{R}_2$\;
    \tcp{\textbf{Select Optimal SQL:} }
    
    $\texttt{SQL}_3 \gets f_\text{LLM}(\mathcal{E}, \mathcal{S'}, \mathcal{V'}, \mathcal{Q}, \mathcal{C}, \mathcal{R}_1, \mathcal{R}_2)$\;
    \Return{$\texttt{SQL}_3$}
}
\BlankLine
\Fn{\texttt{$Step_4: MTSC$}($\mathcal{E}, \mathcal{S'}, \mathcal{V'}, \mathcal{Q}, \mathcal{C}, \texttt{SQL}_3$, N)}{
    Initialize $\texttt{SQL}_4^{(0)} \gets \texttt{SQL}_3$, $i \gets 0$\;
    \While{Not Correct($\texttt{SQL}_4^{(i)}$) and $i < N$}{
        \tcp{\textbf{Capture error or empty results: }}
        
        $E^{(i)} \gets \texttt{CheckError}(\texttt{SQL}_4^{(i)})$\;
        \tcp{\textbf{Refine SQL using LLM:}}
        $\texttt{SQL}_4^{(i+1)} \gets f_\text{LLM}(\mathcal{E}, \mathcal{S'}, \mathcal{V'}, \mathcal{Q}, \mathcal{C}, \texttt{SQL}_4^{(\leq i)}, E^{(\leq i)})$\;
        $i \gets i + 1$\;
    }
    \Return{$\texttt{SQL}_4^{(i)}$}
}

\end{algorithm}

\begin{table*}[htbp]
\centering
\small
\caption{Comparison of Execution Accuracy and Valid Efficiency Score on the BIRD Development Set Across Different Methods and Models. \emph{Open} column indicates whether the code is released or not. \raisebox{1ex}[0pt][0pt]{\scriptsize$*$} symbol denotes fine-tuned methods.}
\label{BIRD_main_resul}

\begin{tabular}{@{}llrccccccc@{}} 
\toprule
\multirow{2}{*}{\textbf{Method}} & \multirow{2}{*}{\textbf{Model}} & \multirow{2}{*}{\textbf{Date}} & \multirow{2}{*}{\textbf{Open}}   & \multirow{2}{*}{\textbf{VES}} & \multicolumn{4}{c}{\textbf{Execution Accuracy}} \\ 
          &        &          &   &   & \textbf{Simple} & \textbf{Moderate} & \textbf{Challenging} & \textbf{Total} \\ \midrule
DIN-SQL~\cite{pourreza2024din}   & GPT-4  &Sep 2023 & \ding{51} & 58.79  & -    & -    & -   & 50.72          \\ 
DAIL-SQL~\cite{gao2023text}    & GPT-4  & Sep 2023 & \ding{51}   & 56.08    & -    & -    & - & 54.76  \\ 
DTS-SQL~\cite{pourreza2024dts}\raisebox{1ex}[0pt][0pt]{\scriptsize$*$}    & DeepSeek-7B &Feb 2024  & \ding{51}    & -   & -  & -     & -   & 55.80           \\ 
Codes~\cite{li2024codes} \raisebox{1ex}[0pt][0pt]{\scriptsize$*$}     & Codes-15B &Feb 2024    & \ding{51}        & 59.87      & -                 & -                    & -                    & 58.47          \\ 
SQL-Palm~\cite{sun2023sql}     & PaLM2   &Mar 2024   & \ding{55}        & -    & 68.92             & 52.07                & 47.89                & 61.93          \\ 
MCS-SQL~\cite{lee2024mcs}     & GPT-4 &May 2024 & \ding{55}        & 64.80       & 70.40              & 53.10                 & 51.40                 & 63.36          \\ 
TA-SQL~\cite{qu2024before}      & GPT-4 &May 2024 & \ding{51}        & -    & 63.14             & 48.60                 & 36.11                & 56.19          \\

CHESS~\cite{talaei2024chess}              & Openllms   &Jun 2024  & \ding{51}        & -    & -                 & -                    & -                    & 61.50           \\
CHESS~\cite{talaei2024chess}              & Proprietary &Jun 2024 & \ding{51}        & -    & -                 & -                    & -                    & 65.00             \\ 
SuperSQL~\cite{li2024dawn}  & GPT-4   &Jul 2024   & \ding{51}        & 61.99      & 66.90              & 46.50                 & 43.80                 & 58.50           \\

MAG-SQL~\cite{xie2024mag}         & GPT-3.5   &Aug 2024   &  \ding{55}       & -    & 65.94             & 46.24                & 40.97                & 57.62          \\

MAG-SQL~\cite{xie2024mag}           & GPT-4  &Aug 2024   &  \ding{55}        & -    & -                 & -                    & -                    & 61.08          \\ 
MAC-SQL~\cite{wang2023mac}         & GPT-3.5   &Sep 2024   & \ding{51}          & -    & -                 & -                    & -                    & 50.56          \\ 
MAC-SQL~\cite{wang2023mac}          & GPT-4 &Sep 2024  & \ding{51}        & 66.39      & 65.73             & 52.69                & 40.28                & 59.39          \\
E-SQL~\cite{caferouglu2024sql}              & GPT-4o-mini  &Sep 2024 & \ding{51}        & -    & 68.00                & 53.23                & 47.59                & 61.60           \\

 E-SQL~\cite{caferouglu2024sql}          & GPT-4o  &Sep 2024     &   \ding{51}       & -    & -                 & -                    & -                    & 65.58 \\ 
 MSc-SQL~\cite{gorti2024msc}\raisebox{1ex}[0pt][0pt]{\scriptsize$*$}    &Gemma-2-9B      &Oct 2024   &\ding{51}  &- &\underline{72.00}  &\textbf{58.00}   &49.00  &\underline{65.60}  \\

\midrule
\textbf{RSL-SQL (ours)}                  & DeepSeek  &Oct 2024   & \ding{51}        & \underline{67.68}             & 69.73             & 54.09                & \textbf{54.48}               & 63.56          \\ 
\textbf{RSL-SQL (ours)}                  & GPT-4o   &Oct 2024    & \ding{51}        & \textbf{70.32}                & \textbf{74.38}              & \underline{57.11}                 & \underline{53.79}                 & \textbf{67.21} \\ \bottomrule
\end{tabular}
\end{table*}

\section{Experiment}

\subsection{Datasets}

\subsubsection{\textbf{BIRD Set}} 
The BIRD dataset \cite{li2024can} is a large-scale, cross-domain Text-to-SQL dataset. Its key characteristic is the emphasis on processing database values while highlighting the challenges posed by dirty external knowledge bases, noisy database values, and the efficiency of SQL queries—especially in large-scale database environments. The SQL queries in BIRD are generally more complex than those found in the Spider dataset.

\subsubsection{\textbf{Spider Set}} The Spider dataset \cite{yu2018spider} is a large-scale, cross-domain Text-to-SQL task dataset. Its key characteristic is that it not only contains complex SQL queries but also covers multi-table databases, making it an important resource for testing model generalization capabilities.
}

\subsubsection{\textbf{Subsampled Development Set (SDS)}}To facilitate ablation studies, reduce computational costs, and preserve the distribution of the BIRD development set, we adopt the Subsampled Development Set outlined in the CHESS\cite{talaei2024chess}. The SDS comprises 10\% of each database in the BIRD development set.

\subsection{Evaluation Metrics}

\subsubsection{\textbf{Execution Accuracy (EX)}}
This metric is defined as the proportion of queries for which the output of the predicted SQL query exactly matches the ground truth SQL query. We report the execution accuracy (EX) as a percentage of the queries in the evaluation set.

\subsubsection{\textbf{Valid Efficiency Score (VES)}}
It measures the efficiency of valid SQL queries generated by models. "Valid SQL queries" are those predicted SQL queries whose result sets match the ground-truth SQL queries.

\subsubsection{\textbf{Non-Strict Recall(NSR)}}
In the existing papers, the recall rate has not been explicitly defined, yet it plays a critical role in determining the accuracy of SQL generation after schema linking. Therefore, we provide a precise definition of recall rate.

We define the \emph{Non-Strict Recall} (NSR) as the ratio of the sum of the number of elements in the intersection between the linked schema set and the ground truth schema set for each question to the total number of elements in the ground truth schema set. Mathematically, this can be expressed as:

\begin{equation}
    \text{NSR} = \frac{\sum_{i=1}^{n} |S_{\text{gt}, i} \cap \tilde{S}_i|}{\sum_{i=1}^{n} |S_{\text{gt}, i}|}
\end{equation}

where \( n \) is the number of questions, \( S_{\text{gt}, i} \) represents the ground truth schema elements for the \( i \)-th question, and \( \tilde{S}_i \) represents the linked schema elements for the \( i \)-th question.

\subsubsection{\textbf{Strict Recall Rate(SRR)}}
If all the necessary columns to generate the correct SQL query are included in \( \tilde{S} \), the value is considered 1; otherwise, it is 0. The \emph{strict recall rate} is the ratio of examples where all required schema elements are successfully recalled to the total number of examples.
\begin{equation}
    \text{SRR} = \frac{\sum_{i=1}^{n} \mathbb{I}(\tilde{S}_i \supseteq S_{\text{gt},i})}{n}
    \label{SRR}
\end{equation}
Our method aims to maximize SRR while keeping $|\tilde{S}|$ as small as possible to reduce prompt size.

Strict Recall Rate is a metric for evaluating the effectiveness of schema linking recall. It is defined as the proportion of correctly recalled column combinations. Specifically, if the column combination recalled by schema linking exactly matches the correct combination, it is considered a successful recall. However, since there is currently only one correct SQL, we first adopt the calculation method described in Equation~\ref{SRR}.

\subsection{Baseline Methods}

In our experiments, we compare our proposed method against a diverse range of state-of-the-art Text-to-SQL approaches. These baselines employ various strategies to enhance performance, representing the current forefront of the field. DIN-SQL\cite{pourreza2024din} and MAC-SQL\cite{wang2023mac} adopt a task decomposition strategy to decompose complex queries into manageable subtasks. E-SQL\cite{caferouglu2024sql} and CHESS\cite{talaei2024chess}, which focus on schema enrichment and linking, aim to address the gap between natural language queries and database structures. SQL-PaLM\cite{sun2023sql} and SuperSQL\cite{li2024dawn} employ distinct prompting and fine-tuning techniques for LLM adaptation in SQL generation. TA-SQL\cite{qu2024before} and CodeS\cite{li2024codes}, which introduce strategies to mitigate hallucinations in LLM-based SQL generation. DAIL-SQL\cite{gao2023text} is designed to address complex database environments. Multi-stage strategies based methods are also included , such as DTS-SQL\cite{pourreza2024dts} that utilizes two-stage fine-tuning,  MAG-SQL\cite{xie2024mag} that adotps multi-agent generative method, and MCS-SQL\cite{lee2024mcs} that leverages multiple prompts and multiple-choice selection. MSc-SQL\cite{gorti2024msc} mitigates the performance gap of smaller open-source models by sampling and comparing multiple SQL query results.

\begin{table}[htbp]

\centering
\small
\caption{Comparison of Execution Accuracy(EX) on Spider Test Set. \raisebox{1ex}[0pt][0pt]{\scriptsize$*$} symbol denotes fine-tuned methods.}
\label{spider_result}
\begin{tabular}{llrc}
\toprule
\textbf{Method}   & \textbf{Model}   &\textbf{Date}    & \textbf{EX}   \\
\midrule
DAIL-SQL~\cite{gao2023text}      & GPT-4   &Sep 2023    & 86.6 \\
DIN-SQL~\cite{pourreza2024din}   & GPT-4  & Sep 2023   & 85.3 \\
DTS-SQL~\cite{pourreza2024dts}\raisebox{1ex}[0pt][0pt]{\scriptsize$*$}   & DeepSeek 7B  &Feb 2024 & 84.4 \\
MCS-SQL~\cite{lee2024mcs}   & GPT-4    &May 2024   & \textbf{89.6} \\
TA-SQL~\cite{qu2024before}      &GPT-4      &May 2024   &85.0   \\
CHESS~\cite{talaei2024chess}  &Openllms      &Jun 2024   &87.2   \\
PET-SQL~\cite{talaei2024chess}  & GPT-4   &Jun 2024    & 87.6 \\
MAG-SQL~\cite{xie2024mag}  &GPT-4     &Aug 2024       &85.6   \\
MAC-SQL~\cite{wang2023mac}   & GPT-3.5-Turbo  &Sep 2024   & 75.5 \\
MAC-SQL~\cite{wang2023mac}  & GPT-4  & Sep 2024    & 82.8 \\
MSc-SQL~\cite{gorti2024msc}\raisebox{1ex}[0pt][0pt]{\scriptsize$*$} &Gemma-2-9B &Oct 2024   &84.7   \\
\midrule
\textbf{RSL-SQL (ours)} &DeepSeek  &Oct 2024    &87.5\\
\textbf{RSL-SQL (ours)}  &GPT-4o    &Oct 2024       &\underline{87.9}   \\
\bottomrule
\end{tabular}

\end{table}

\section{Results and Analysis}

\subsection{Main Results}

\subsubsection{BIRD Results}We conduct experiments on the BIRD development set using the GPT-4o and DeepSeek models, demonstrating the performance of our proposed RSL-SQL framework. We compare it with 17 baselines, and the results are shown in Table~\ref{BIRD_main_resul}. We conduct a comparative analysis of fine-tuning-based methods and LLM-based methods. The MSc-SQL baseline currently holds the SOTA metric among fine-tuning-based approaches on the BIRD development set. However, its execution accuracy remains slightly lower than our framework utilizing the GPT-4o model. Notably, several baselines shown in Table~\ref{BIRD_main_resul} leveraging the GPT-4 model fail to match the performance of our framework using the DeepSeek model. Under the same condition of employing the GPT-4o model, our approach not only achieves higher execution accuracy but also incurs significantly lower costs compared to the E-SQL method. With an execution accuracy of 67.21\%, our framework sets a new SOTA benchmark in the open-source domain.

\subsubsection{Spider Results}In order to evaluate the generalizability of the proposed RSL-SQL, we further conduct experiments on the Spider test set, and the results are shown in Table~\ref{spider_result}. RSL-SQL achieves an execution accuracy of 87.7\% when using the DeepSeek model, which improves to 87.9\% with the GPT-4o model. This performance aligns closely with the latest MCS-SQL model (GPT-4), which achieves an execution accuracy of 89.6\%. This highlights the strong generalizability of RSL-SQL and its potential for generating high-quality Text-to-SQL.

\begin{table}[htbp]
\centering
\caption{Comparison of Schema Recall Rate (SRR) and Non-Strict Recall (NSR) metrics across different methods on the BIRD dev set, with Avg. T and Avg. C representing the average number of tables and columns input per question, respectively. DeepSeek is abbreviated as (D) and GPT-4o as (G).}
\label{schema_linking}
\small
\begin{tabular}{lcccc}
    \toprule
    \textbf{Method}        & \textbf{NSR} & \textbf{SRR}    &\textbf{Avg. T}   &\textbf{Avg. C} \\ 
    \midrule
    Full Schema                  & 100            &  100   &  7.44 &  76.28   \\
    Gold-Based                  & 100            & 100    &  1.94 &  4.74   \\
    \midrule
    HySCSL~\cite{maamari2024death}                   & -            & 90.36    &  - &  -   \\
    SCSL~\cite{maamari2024death}                      & -            & 88.77   & -   & -    \\
    HyTCSL~\cite{maamari2024death}                   & -            & 83.00    & -    & -  \\
    TCSL~\cite{maamari2024death}                     & -            & 77.44   & -  &  -    \\
    MCS\cite{lee2024mcs}   & -        &  89.80        & -  &  -  \\
    CHESS\cite{talaei2024chess}                     & 94.00        & 89.70      &  \textbf{1.92} &   \textbf{4.47}  \\
    \midrule
    RSL-SQL(D)& \underline{97.29}  &\underline{93.28} &  \underline{5.36} &  14.85   \\
    \hspace{1em} $\textbf{·} $ Forward &87.55 &78.81 &3.45 &10.98 \\
    \hspace{1em} $\textbf{·} $ Backward &94.39 &88.46 &5.26 &10.13 \\  
    \midrule
    RSL-SQL(G)    & \textbf{97.69} & \textbf{94.32}   & 5.40  & \underline{13.02}   \\ 
    \hspace{1em} $\textbf{·} $ Forward &90.04 &84.74 &3.43 &9.12 \\
    \hspace{1em} $\textbf{·} $ Backward &95.54 &90.48 &5.31 &10.29 \\
    \bottomrule
\end{tabular}

\end{table}

\subsubsection{Schema Linking Results}

The quality of the schema link affects the accuracy of the final generated SQL and also affects the length of the input token. To evaluate the effect of our proposed bidirectional schema linking, we conduct experiments on the BIRD dataset and report the recall metrics as reported in previous studies. 

As shown in Table~\ref{schema_linking}, the CHESS method achieves schema linking through retrieval combined with multi-round filtering using LLMs, attaining a strict recall rate of 89.7\%. The MCS-SQL approach employs multi-prompt and multi-choice decoding strategies, leveraging LLMs for iterative selection to achieve a strict recall rate of 89.8\%. In contrast, our method requires only one to two input rounds, significantly reducing token consumption. Utilizing the GPT-4o model, our bidirectional schema linking approach achieves SOTA performance on both NSR and SRR metrics, fully recalling 94\% of the columns required for SQL generation.

In terms of performance, our forward schema linking demonstrates a relatively lower recall rate, while the backward schema linking achieves nearly 90\% recall, highlighting its critical importance. Notably, among previous methods, only PET-SQL applied a similar strategy on the Spider test dataset, but it focused solely on table simplification without addressing column simplification~\cite{li2024pet}. Our approach not only recalls the vast majority of tables but also significantly reduces the number of columns. Furthermore, due to the highly complementary nature of forward and backward schema linking, the union of columns from both methods does not substantially increase the total number of columns. As a result, our bidirectional schema linking achieves SOTA performance overall.

\begin{table*}[htbp]

\centering
\small
\caption{Execution Accuracy Across Different Difficulty Levels on the BIRD Dev Set and the Contribution of Each Component.}

\label{step_by_step}
\begin{tabular}{llccccl} \toprule
\multirow{2}{*}{\textbf{Model}}  & \multirow{2}{*}{\textbf{Step}} & \multicolumn{4}{c}{\textbf{Execution Accuracy}} \\  
             &      & \textbf{Simple} & \textbf{Moderate} & \textbf{Challenging} & \textbf{Total}  \\ \midrule
\multirow{7}{*}{DeepSeek} 
        &Base prompt    &49.84  &34.70  &25.52  &42.96  \\
        &\hspace{1em}+ Few-shot examples    &63.24  &45.04  &39.31  &$55.48_{\textcolor{myred}{\uparrow12.52}}$ \\
        &\hspace{1em}+ Data samples &66.05  &46.77  &38.62  &$57.63_{\textcolor{myred}{\uparrow2.15}}$ \\   
        &  \textbf{Step 1:} Bidirectional Schema Linking   & 66.38 & 47.41 & 46.90 & $58.80_{\textcolor{myred}{\uparrow1.17}}$  \\
       
       &  \textbf{Step 2:} Contextual Information Augmentation   & 68.76 & 51.94 & 47.59 & $61.67_{\textcolor{myred}{\uparrow2.87}}$  \\    
       &  \textbf{Step 3:} Binary Selection Strategy     & 69.73 & 53.88 & 52.41 & $63.30_{\textcolor{myred}{\uparrow1.63}}$  \\
       &  \textbf{Step 4:} Multi-Turn Self-Correction      & \textbf{69.73} & \textbf{54.09} & \textbf{54.48} & \textbf{$\textbf{63.56}_{\textcolor{myred}{\uparrow0.26}}$}  \\
 \midrule
\multirow{7}{*}{GPT-4o  } 
        &Base prompt    &57.73  &37.07  &28.28  &48.70  \\
        &\hspace{1em}+ Few-shot examples &67.35 &46.34  &42.76  &$58.67_{\textcolor{myred}{\uparrow9.97}}$   \\
        &\hspace{1em}+ Data samples &68.86  &51.29  &53.10  &$62.06_{\textcolor{myred}{\uparrow3.39}}$  \\ 
        & \textbf{Step 1:} Bidirectional Schema Linking    & 70.05 & 52.59 & 48.28 & $62.71_{\textcolor{myred}{\uparrow0.65}}$  \\
     
      &  \textbf{Step 2:} Contextual Information Augmentation   & 72.11 & 54.74 & 53.10 & $65.06_{\textcolor{myred}{\uparrow2.35}}$  \\
      &  \textbf{Step 3:} Binary Selection Strategy     & 73.51 & 56.47 & 53.10 & $66.43_{\textcolor{myred}{\uparrow1.37}}$  \\
      &  \textbf{Step 4:} Multi-Turn Self-Correction     & \textbf{74.38} & \textbf{57.11} & \textbf{53.79} & \textbf{$\textbf{67.21}_{\textcolor{myred}{\uparrow0.78}}$}  \\
 \bottomrule
\end{tabular}

\end{table*}

\begin{figure*}[htbp]
    \centering

    \begin{subfigure}[t]{0.45\textwidth}  
        \centering
        \includegraphics[width=\linewidth]{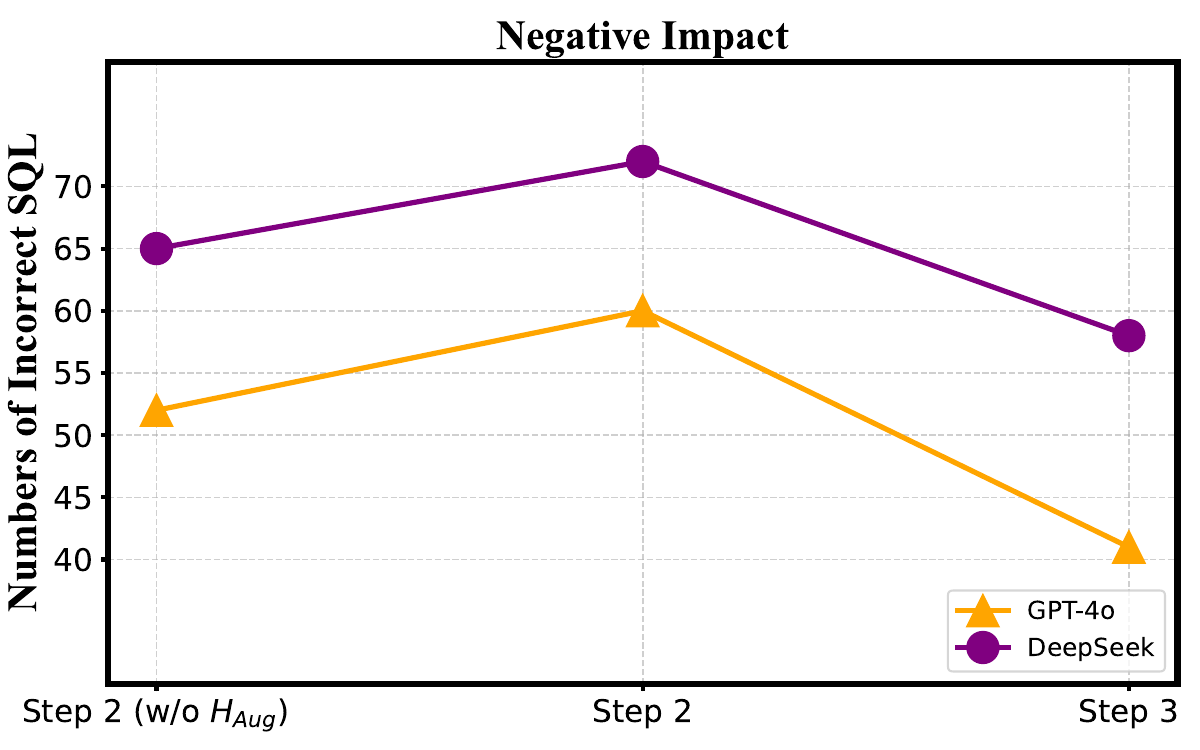}
        \caption{Negative Impact: SQLs initially correct become incorrect under \textbf{Step 2 (w/o $H_\text{Aug}$)}, \textbf{Step 2}, and \textbf{Step 3}.}
        \label{fig:left_image}
    \end{subfigure}%
    \hfill  
    \begin{subfigure}[t]{0.45\textwidth}  
        \centering
        \includegraphics[width=\linewidth]{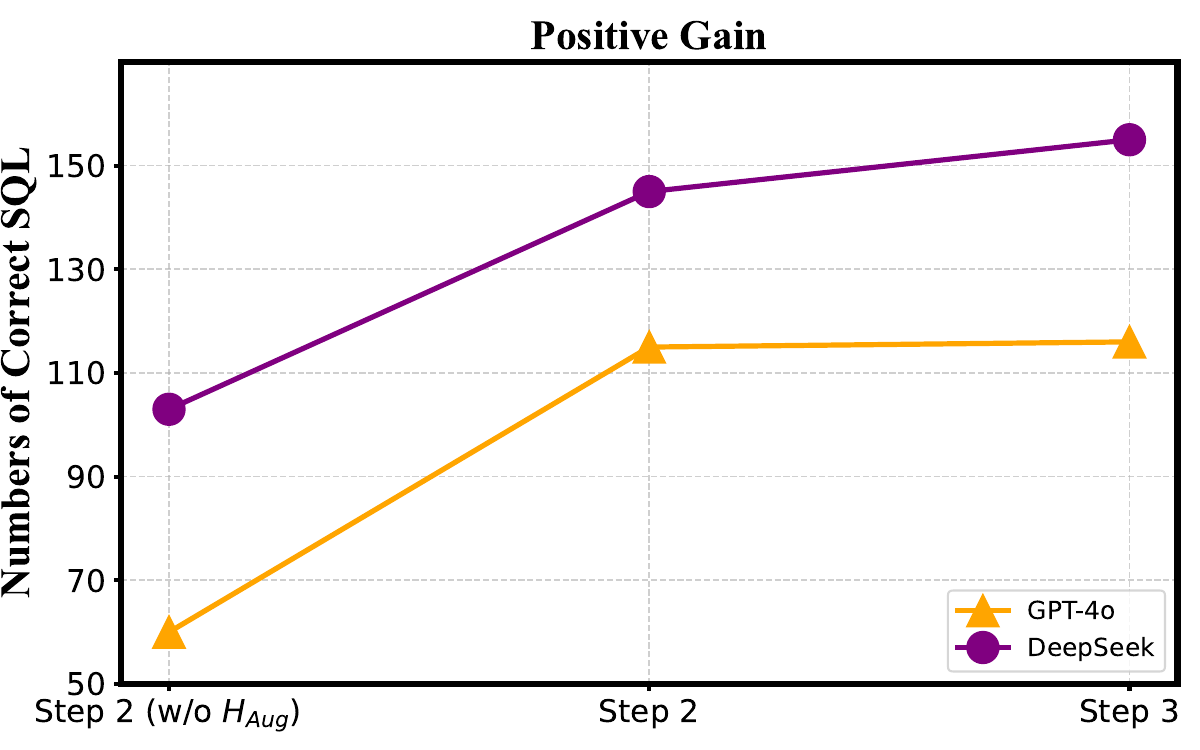}
        \caption{Positive Gain: SQLs initially incorrect become correct under \textbf{Step 2 (w/o $H_\text{Aug}$)}, \textbf{Step 2}, and \textbf{Step 3}.}
        \label{fig:right_image}
    \end{subfigure}

    \caption{Negative Impact and Positive Gain.}
    \label{fig:overall}
\end{figure*}

 \begin{table}[htbp]
\caption{Token Consumption and Cost Analysis Across Different Methods.}\label{cost}
\small
\centering
\scalebox{0.8}{
\begin{tabular}{llccll}
\toprule
\textbf{Method} & \textbf{Model}    & \textbf{Input(M)}   & \textbf{Output(M)} & \textbf{Cost(\$)}     & \textbf{EX}  \\
\midrule
MAC-SQL    & GPT-4    & 9.63   & 0.89  & 342.44  & 59.39 \\
TA-SQL     & GPT-4    & 10.71 & 0.51  & 351.55  & 56.19 \\
E-SQL  & GPT-4o   & 67.28  & 1.44  & 182.56  & 65.58 \\
\midrule
\textbf{RSL-SQL (ours)}    & DeepSeek & 21.90  & 0.73  & 3.27    & 63.56 \\
\textbf{RSL-SQL (ours)}    & GPT-4o   & 21.90  & 0.73  & 62.06   & 67.21    \\
\bottomrule
\end{tabular}}

\end{table}

\begin{table*}[htbp]

\centering
\small
\caption{Impact of Information Augmentation Components in Step 2 on Execution Accuracy.}

\label{ablation_step2}
\begin{tabular}{llccccl} \toprule
\multirow{2}{*}{\textbf{Model}} & \multirow{2}{*}{\textbf{Step}}  & \multicolumn{4}{c}{\textbf{Execution Accuracy}} \\  
                &  & \textbf{Simple} & \textbf{Moderate} & \textbf{Challenging} & \textbf{Total}  \\ \midrule
\multirow{4}{*}{DeepSeek} 
       &  \textbf{Step 2:} Contextual Information Augmentation   & 68.76 & 51.94 & 47.59 & \textbf{61.67}  \\
       & \hspace{1em} - \textbf{w/o} Description of Columns        & 67.89  & 50.00       & 44.83       & $60.30_{\textcolor{mydarkgreen}{\downarrow1.37}}$  \\
        & \hspace{1em} - \textbf{w/o} SQL Components Generation               & 67.78     & 53.66        & 43.45           & $61.21_{\textcolor{mydarkgreen}{\downarrow0.46}}$     \\

        & \hspace{1em} - \textbf{w/o} Both               & 66.70   & 49.57    & 47.59       & $59.71_{\textcolor{mydarkgreen}{\downarrow1.96}}$ \\ \midrule
\multirow{4}{*}{GPT-4o  }  
      & \textbf{Step 2:} Contextual Information Augmentation   & 72.11 & 54.74 & 53.10 & \textbf{65.06}  \\
      &\hspace{1em} - \textbf{w/o} Description of Columns        & 71.03  & 53.66    & 51.03       & $63.89_{\textcolor{mydarkgreen}{\downarrow1.17}}$ \\
    & \hspace{1em} - \textbf{w/o} SQL Components Generation               & 71.24      & 54.09        & 48.97           & $63.95_{\textcolor{mydarkgreen}{\downarrow1.11}}$     \\
    & \hspace{1em} - \textbf{w/o} Both               & 70.70   & 49.35    & 49.66       & $62.26_{\textcolor{mydarkgreen}{\downarrow2.80}}$  \\ \bottomrule
\end{tabular}
\end{table*}

\subsection{Analysis}

We conduct an ablation study to investigate the incremental impact of each component of our proposed method on execution accuracy (EX). The results of this study on the BIRD development set are presented in Table~\ref{step_by_step}. This table illustrates the execution accuracy of DeepSeek and GPT-4o across various task difficulty levels, showing the effects achieved by gradually incorporating different experimental steps into the models.

\subsubsection{Prompt Refinement}
For large language models, the prompt plays a crucial role. Before conducting our experiments, we refine the prompt by adjusting its structure and the information it provides. As shown in Table~\ref{step_by_step}, we validate the effectiveness of each piece of additional information included in the prompt.

The \textbf{Basic Prompt} is built upon a complete database schema, including table names, column names, and foreign key information. It also includes user questions and additional context. The addition of \textbf{few-shot examples} results in a 10\% improvement in execution accuracy for both DeepSeek and GPT-4o, highlighting the importance of these examples. Following this, the inclusion of additional \textbf{data samples} contributes to a further 2\% increase in accuracy, marking the completion of our prompt adjustment process. After fine-tuning the prompt, even without adding any additional components, the execution accuracies reach 62.06\% and 57.63\% when using the GPT-4o and DeepSeek models, respectively. This further demonstrates the robustness of our prompt.

\subsubsection{RSL-SQL Framework}

As shown in Table~\ref{step_by_step}, each step of our framework effectively improves the execution accuracy, with increases ranging from 1\% to 3\%, ultimately helping the framework achieve open-source SOTA performance. Among these steps, Step 2 and Step 3 are the core of our approach, contributing over 70\% of the overall improvement.

\paragraph{ \emph{\textbf{Step 1: BSL}}}~{     Building on the previous prompt, we incorporate the results from \textbf{Forward Schema Linking} into the prompt to generate preliminary SQL queries. This integration leads to an approximate 1\% performance improvement. This preliminary SQL is generated within a complete database schema, ensuring the integrity of the database structure. At this step, we also use the bidirectional schema linking method to simplify the database schema, aiming to recall the necessary columns as much as possible, thereby reducing input noise in the subsequent steps.}

 \paragraph{\emph{\textbf{Step 2: CIA}}}~{As shown in Table~\ref{step_by_step}, this step is one of the core steps of our approach, contributing to a 2\% to 3\% improvement in our execution accuracy. As illustrated in Fig.~\ref{fig:left_image}, schema linking can sometimes introduce errors by turning previously correct SQL queries on the full schema into incorrect ones, demonstrating a potential negative impact of schema linking. Contextual information augmentation may slightly increase this negative impact; however, as shown in Fig.~\ref{fig:right_image}, it significantly enhances the positive gain of schema linking by greatly increasing the number of corrected queries. Through a sampled inspection of 20 examples, we find that the generated textual information effectively aids the model in understanding the mapping between user queries and the database, as well as in identifying keyword matches. This is the reason why this step significantly enhances the positive gain. Since the positive gain far outweighs the negative, contextual information augmentation effectively mitigates the risks of schema linking.} 

 \paragraph{\emph{\textbf{Step 3: BSS}}}~{The effectiveness of Binary Selection Strategy lies in the fact that complete database structure information enables the LLM to gain a comprehensive understanding of the overall database. However, this can also introduce information redundancy and interference. In contrast, a simplified database structure reduces such interference but may lead to incomplete column recall, thus compromising the integrity of the database structure. Therefore, SQL generated by the two approaches has its own strengths and weaknesses. 
 
 As shown in Table~\ref{step_by_step},  this step plays a pivotal role in our approach, driving a 1.5\% enhancement in execution accuracy. The purpose of this step is to mitigate the risks associated with schema linking, reducing its negative impact while increasing the positive gain. As shown in Fig.~\ref{fig:left_image}, after applying the binary selection strategy, the negative impact is significantly reduced, regardless of whether the DeepSeek or GPT-4o model is used. Fig.~\ref{fig:right_image}  illustrates that using the DeepSeek model significantly increases the positive gain, while the GPT-4o model provides a slight increase in positive gain. This is because, after \emph{Step 2: CIA} with the GPT-4o model, the execution accuracy has already reached 65.06\%, and the potential for further performance improvement through the selection strategy becomes limited.}

\begin{figure*}[htbp]
    \centering

    \begin{subfigure}[t]{0.45\textwidth}  
        \centering
        \includegraphics[width=\linewidth]{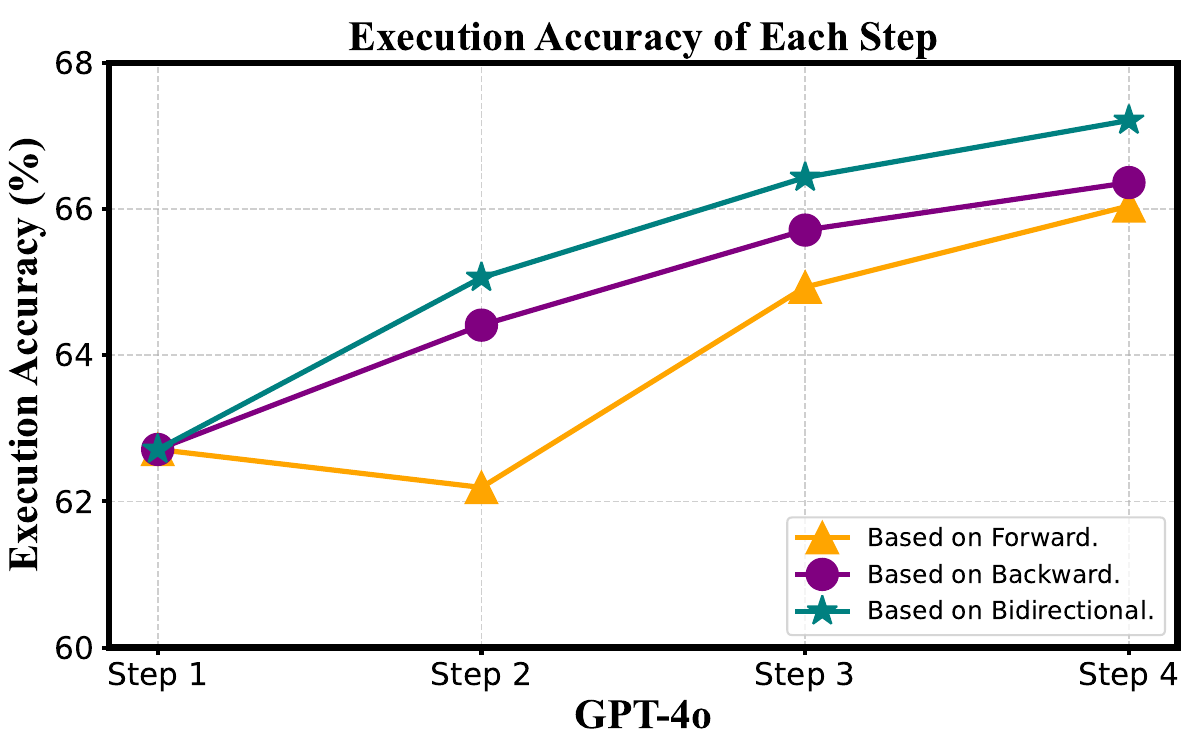}
        \caption{Execution Accuracy of Each Step Using the GPT-4o.}
        \label{fig:left_risk1}
    \end{subfigure}%
    \hfill  
    \begin{subfigure}[t]{0.45\textwidth}  
        \centering
        \includegraphics[width=\linewidth]{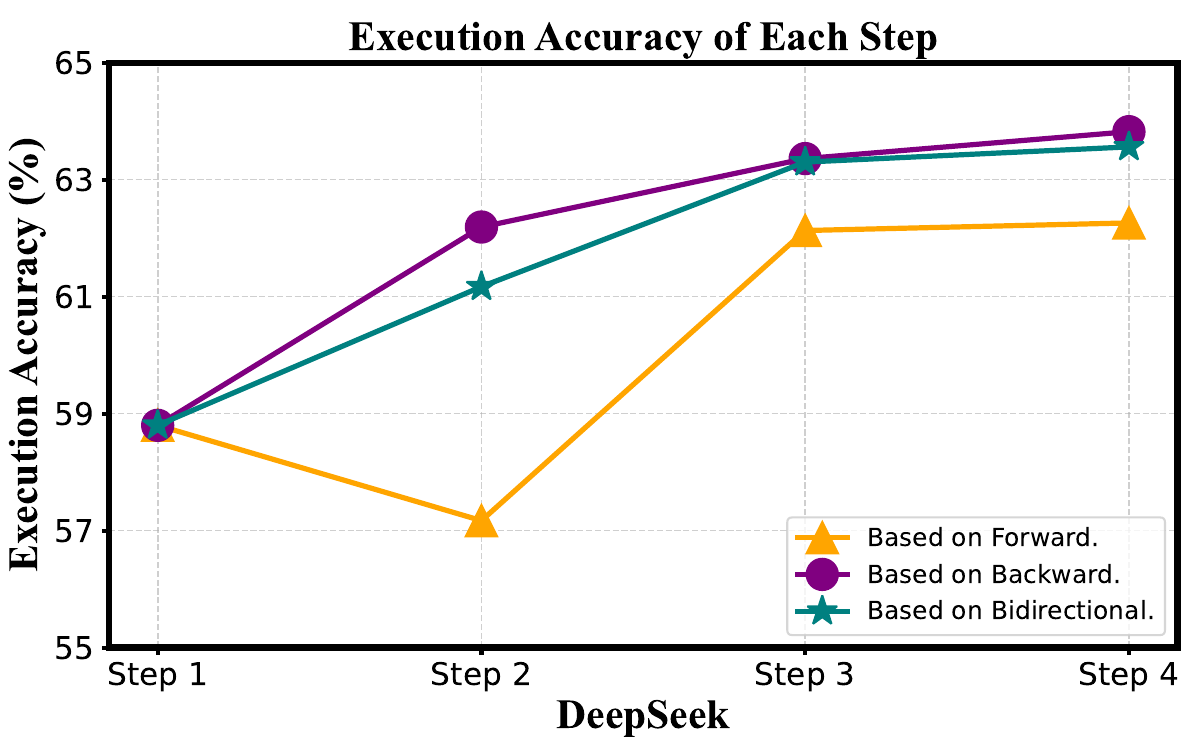}
        \caption{Execution Accuracy of Each Step Using the DeepSeek.}
        \label{fig:right_risk1}
    \end{subfigure}


    \caption{On the BIRD development set, schema linking based on  the results of forward  (\textbf{Forward.}), backward  (\textbf{Backward.}), and bidirectional schema linking (\textbf{Bidirectional}) is performed, and the execution accuracy for each step is presented.}
    \label{fig:risk1}
\end{figure*}

 \paragraph{\emph{\textbf{step 4: MTSC}}}~{To further enhance the accuracy of generated SQL statements, we propose this strategy. When SQL fails to execute or returns empty results, it often indicates an issue with the query. By using rules to assess the execution risks of SQL, we can regenerate and adjust high-risk SQL queries. As shown in Table~\ref{step_by_step}, this strategy slightly improves the accuracy of SQL generation.}

\subsubsection{Token Consumption and Cost}~{In order to verify the low consumption of our method, we conduct experiments on the MAC-SQL, TA-SQL, and E-SQL methods and estimated their token consumption and cost. The results are shown in Table~\ref{cost}.

The comparison results show that RSL-SQL with DeepSeek model has lower overhead and performs better than some methods using the GPT-4 model. When using the GPT-4o model, the E-SQL method consumes three times as many tokens as ours, with performance about 2\% lower.

}

\subsection{Ablation Study}

\subsubsection{Bidirectional Schema Linking}

Our bidirectional schema linking consists of two components: forward schema linking and backward schema linking. To evaluate the impact of using only forward or backward schema linking on subsequent steps, we conducted ablation experiments. The results are shown in Fig.~\ref{fig:risk1}.

As shown in Fig.~\ref{fig:left_risk1}, when using the GPT-4o model, the execution accuracy achieved in subsequent steps with bidirectional schema linking is higher than that achieved with either forward or backward schema linking alone. However, Fig.~\ref{fig:right_risk1} indicates that for the DeepSeek model, the execution accuracy in subsequent steps is slightly higher when only backward schema linking is used compared to bidirectional schema linking. This occurs because, for the DeepSeek model, while bidirectional schema linking achieves higher recall, it also introduces more noise, which negatively affects SQL generation. This phenomenon aligns with the findings of Paper~\cite{maamari2024death}: for stronger models (e.g., GPT-4o), precision has a lesser impact on execution accuracy, and improved recall typically leads to higher execution accuracy. In contrast, for weaker models (e.g., DeepSeek), precision has a more pronounced effect, and excessively high recall does not necessarily translate into higher execution accuracy.

As illustrated in Fig.~\ref{fig:risk1}, when using either the GPT-4o or DeepSeek model, the execution accuracy in \emph{Step 2: CIA} based solely on forward schema linking results shows a downward trend. This is due to the lower recall rate of forward schema linking, further underscoring the necessity of backward schema linking in the schema linking process. 

However, even under these circumstances, the execution accuracy based on forward schema linking results significantly improves after applying our binary selection strategy, demonstrating the strategy’s effectiveness and robustness. Although the results of forward schema linking are suboptimal and may introduce some noise, its role in enhancing recall is indispensable, laying a solid foundation for subsequent schema linking optimizations.

\begin{table}[]
\centering
\caption{Execution accuracy (EX) on Subsampled Development Set, $H_\text{EKC} = \{ H_\text{E}, H_\text{C}, H_\text{K}\}$.}\label{HKEC}
\scalebox{0.75}{
\begin{tabular}{clcccc}
\toprule
\multicolumn{1}{l}{\multirow{2}{*}{\textbf{Model}}} & \multirow{2}{*}{\textbf{Setting}}           & \multicolumn{4}{c}{\textbf{Execution Accuracy}}                  \\
\multicolumn{1}{l}{}                       &                                    & \textbf{Simple} & \textbf{Moderate} & \textbf{Challenging} & \textbf{Total} \\
\midrule
\multirow{5}{*}{DeepSeek}                  & Simplified Schema + $H_\text{EKC}$ & 69.14  & 38.89    & 50.00       & 56.46 \\
                                           & \hspace{1em} - w/o $H_\text{E}$                          & 62.96  & 46.30    & 50.00       & $55.78_{\textcolor{mydarkgreen}{\downarrow0.68}}$ \\
                                           & \hspace{1em} - w/o $H_\text{K}$                             & 64.20  & 46.30    & 50.00       & $56.46_{\textcolor{mydarkgreen}{\downarrow0.00}}$ \\
                                           & \hspace{1em} - w/o $H_\text{C}$                           & 64.20  & 42.59    & 50.00       & $55.10_{\textcolor{mydarkgreen}{\downarrow1.36}}$ \\
                                           & \hspace{1em} - w/o $H_\text{EKC}$                            & 61.73  & 44.44    & 50.00       & $54.42_{\textcolor{mydarkgreen}{\downarrow2.04}}$ \\
                                        \midrule
\multirow{5}{*}{GPT-4o}                    & Simplified Schema + $H_\text{EKC}$ & 72.84  & 51.85    & 41.67       & 62.59 \\
                                           & \hspace{1em} - w/o $H_\text{E}$                          & 71.60  & 44.40    & 50.00       & $59.86_{\textcolor{mydarkgreen}{\downarrow2.73}}$ \\
                                           & \hspace{1em} - w/o $H_\text{K}$                             & 72.84  & 50.00    & 41.67       & $61.90_{\textcolor{mydarkgreen}{\downarrow0.69}}$ \\
                                           & \hspace{1em} - w/o $H_\text{C}$                            & 70.37  & 50.00    & 50.00       & $61.22_{\textcolor{mydarkgreen}{\downarrow1.37}}$ \\
                                           & \hspace{1em} - w/o $H_\text{EKC}$                            & 69.14  & 50.00    & 50.00       & $60.54_{\textcolor{mydarkgreen}{\downarrow2.05}}$   \\
                                           \bottomrule
\end{tabular}}
\end{table}

\subsubsection{Contextual Information Augmentation}

\emph{Step 2: CIA} is one of the core steps of our RSL-SQL framework and significantly increases the positive gain. The key function of this step lies in its components, where the generated textual information helps the LLM understand the mapping between the user query and the database, as well as the usage of keywords. As shown in Table~\ref{ablation_step2}, these components contribute a performance improvement of about 2\% to 3\%. Whether using DeepSeek or GPT-4o, the description of columns shows an improvement of over 1\%. However, for SQL components, the improvement with DeepSeek is only 0.46\%. Upon sampling a subset of examples, we found that the lower quality of the component text generated by DeepSeek resulted in a less significant performance gain.

In order to further verify the effectiveness of each component in \textbf{SQL Components Generation}, we conduct ablation experiments on the Subsampled Development Set. As shown in Table~\ref{HKEC}, each component of SQL Components Generation significantly contributes to the execution accuracy, whether on DeepSeek or GPT-4o models. In addition, SQL Components Generation can improve the execution accuracy by more than 2\% as a whole, which fully proves its effectiveness in model performance optimization.

\section{Case Study}

\begin{figure*}[htbp]
    \centering
    \includegraphics[width=1\linewidth]{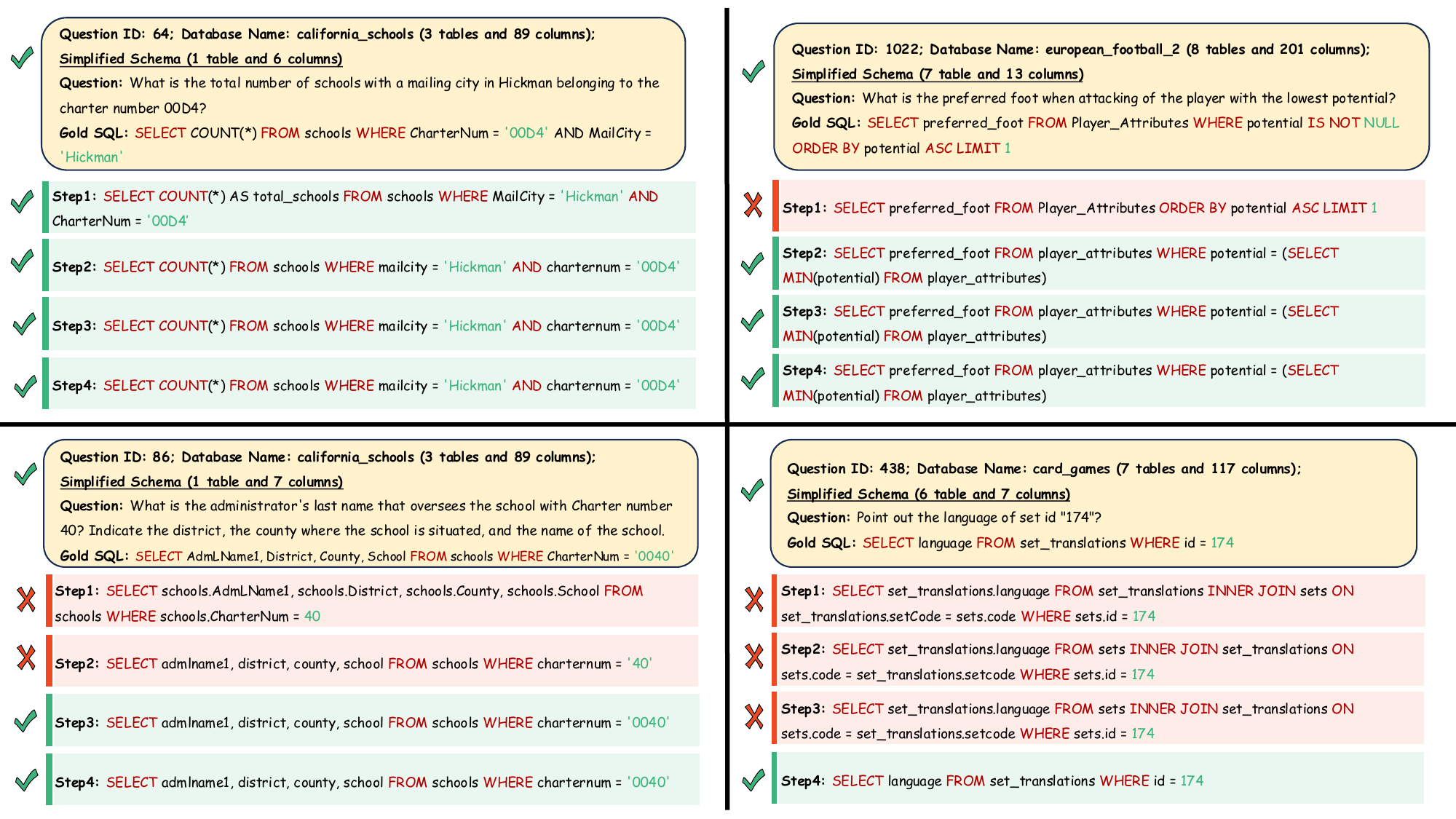}
    \caption{Four Examples Demonstrating the Robustness of Our Framework.}
    \label{case_study}
\end{figure*}

The RSL-SQL framework ensures the correctness of SQL generation through contextual information augmentation, a binary selection strategy, and multi-turn self-correction. As shown in Fig.~\ref{case_study}, even if errors occur in earlier steps, the subsequent steps can correct these errors and improve SQL accuracy through specific methods.

The core of the framework lies in \emph{Step 2: CIA} and \emph{Step 3: BSS}. To illustrate more intuitively how contextual information augmentation enhances performance by increasing positive gain and how binary selection strategy mitigates risks by reducing negative impact, we have selected two representative examples to demonstrate their underlying mechanisms in detail.

\subsection{Maximize Positive Gain}

The description of each column is undoubtedly useful because it helps LLM understand the specific meaning of each column and build a mapping relationship between user questions and columns. The textual information generated by LLM covers database elements, SQL keywords, and query conditions. Among them, the further filtered database elements can guide LLM to focus on elements that are more likely to be used; the generated SQL keywords can remind LLM of critical but often overlooked constraints; the generated conditions are obtained from the user What is obtained by decomposing the problem can help LLM solve the problem step by step in a more orderly manner. With this enhanced information, LLM is able to focus more attention on key parts that were previously under-focused, thereby significantly increasing positive gain. For example, the following is an example in BIRD,

\noindent \textbf{Question:}

\emph{What is the preferred foot when attacking of the player with }

\emph{the lowest potential?}




\noindent \textbf{Contextual Information:}

\texttt{Elements: }

\texttt{[player\_attributes.\textcolor{mydarkgreen}{potential}]}

\texttt{[player\_attributes.\textcolor{mydarkgreen}{preferred\_foot}]}

\texttt{Key Words: [\textcolor{mydarkgreen}{\textbf{'MIN'}}, '=']}

\texttt{Conditions: }

\texttt{['preferred foot when attacking']}

\texttt{['player with the \textcolor{mydarkgreen}{lowest potential}']}

\noindent \textbf{\ding{55} SQL$_1$ Genereted in Step 1:}

\texttt{\textcolor{myred}{\textbf{SELECT}} preferred\_foot }

\texttt{\textcolor{myred}{\textbf{FROM}} Player\_Attributes }

\texttt{\textcolor{myred}{\textbf{ORDER BY}} potential \textcolor{myred}{ASC} }

\texttt{\textcolor{myred}{\textbf{LIMIT}} 1 }

\noindent \textbf{\ding{51} SQL$_2$ Generated in Step 2:}

\texttt{\textcolor{myred}{\textbf{SELECT}} preferred\_foot }

\texttt{\textcolor{myred}{\textbf{FROM}} player\_attributes }

\texttt{\textcolor{myred}{\textbf{WHERE}} potential = (}

\texttt{   \textcolor{myred}{\textbf{SELECT}} \textcolor{myred}{MIN}(potential)  }

\texttt{   \textcolor{myred}{\textbf{FROM}} player\_attributes}

\texttt{)}

\noindent In this example, the incorrect generation of SQL$_1$ stems from failing to account for parallel conditions, leading to an initial SQL with deviations. The generated textual information comprehensively covered the critical columns required for SQL generation, and the inclusion of the keyword \textcolor{mydarkgreen}{\textbf{\texttt{`MIN'}}} explicitly indicated the possibility of parallel conditions. This supplementary information enabled the LLM to better grasp the semantic nuances of the query and the structure of the database, effectively guiding it to produce the correct SQL.

\subsection{Minimize Negative Impact}

The binary selection strategy is a pivotal step in our approach. By choosing the more accurate SQL from the two statements generated using the complete schema and the simplified schema, respectively, this strategy strikes a balance between preserving schema completeness and minimizing noise. An illustrative example is provided below,

\noindent \textbf{Question:}

\emph{Is molecule TR151 carcinogenic? }

\noindent \textbf{\ding{51} SQL$_1$ Generated in Step 1:}

\texttt{\textcolor{myred}{\textbf{SELECT}} label }

\texttt{\textcolor{myred}{\textbf{FROM}} molecule }

\texttt{\textcolor{myred}{\textbf{WHERE}} molecule\_id = 'TR151'}

\noindent \textbf{\ding{55} SQL$_2$ Generated in Step 2:}

\texttt{\textcolor{myred}{\textbf{SELECT}} label }

\texttt{\textcolor{myred}{\textbf{FROM}} molecule }

\texttt{\textcolor{myred}{\textbf{WHERE}} molecule\_id = 'TR151' }

\texttt{\textcolor{myred}{\textbf{AND}} \textcolor{mydarkgreen}{label = '+' }}

\noindent \textbf{SQL execution results:}

\texttt{SQL$_1$ execution results:}

\texttt{[Row\_count, Column\_count] = [1, 1]}

\texttt{[Result] = \textcolor{mydarkgreen}{[('-',)]}}

\texttt{SQL$_2$ execution results:}

\texttt{[Row\_count, Column\_count] = [0, 0]}

\texttt{[Result] = \textcolor{mydarkgreen}{[]}}

\noindent \textbf{\ding{51} SQL$_3$ Selected in Step 3:}

\texttt{\textcolor{myred}{\textbf{SELECT}} label }

\texttt{\textcolor{myred}{\textbf{FROM}} molecule }

\texttt{\textcolor{myred}{\textbf{WHERE}} molecule\_id = 'TR151'}

\noindent Introducing new components in Step 2 may inevitably lead to some originally correct SQL statements being modified incorrectly. However, our binary selection strategy minimizes this negative impact as much as possible. By incorporating execution information, the strategy enables the LLM to further analyze the candidate results, facilitating a more accurate selection of the correct SQL and significantly reducing associated risks.

\section{Conclusion}

We propose the RSL-SQL framework, which achieves open-source state-of-the-art performance on the BIRD development set. The framework employs a bidirectional schema linking approach, attaining a strict recall rate of 94\%. We conduct an in-depth analysis of how schema linking recall rates affect the execution accuracy of SQL generation for different models. Through ablation studies, we rigorously validate the effectiveness of each component within the framework and thoroughly investigate their operational mechanisms. Additionally, we compare the computational costs of various methods, demonstrating that our framework achieves an excellent balance between performance and cost, making it a highly efficient and cost-effective solution.




\bibliographystyle{IEEEtran}
\bibliography{icde}


\end{document}